    \documentclass[12pt,psf,epsf]{article}
\usepackage[centertags]{amsmath}
\usepackage{latexsym}
\usepackage{mathtools}
\usepackage{amssymb}
\usepackage{graphicx}
\usepackage{epsfig}
\usepackage{ulem}
\usepackage[english]{babel}
\usepackage{array}
\usepackage{amsthm}
\usepackage{latexsym}
\usepackage[mathcal]{euscript}
\pdfoutput=1
\usepackage{epsfig}

 \usepackage{jheppub}
 \usepackage{hyperref}

\usepackage{subfig}
\usepackage{psfrag}

\usepackage[utf8]{inputenc}
\usepackage{float}
\usepackage{cancel}
\usepackage{mathrsfs}
\usepackage{amssymb}
\usepackage{amsfonts}
\usepackage{amsmath}
\usepackage{slashed}
\usepackage{bm}
\usepackage{color}
\usepackage{appendix}
\newcommand{\be}{\begin{equation}}
\newcommand{\ee}{\end{equation}}
\newcommand{\bea}{\begin{eqnarray}}
\newcommand{\eea}{\end{eqnarray}}
\newcommand{\bwt}{\begin{widetext}}
\newcommand{\ewt}{\end{widetext}}

\newcommand{\bi}{\begin{itemize}}
\newcommand{\ei}{\end{itemize}}

\usepackage{setspace}

\definecolor{dgreen}{rgb}{0.,0.6,0.}

\newcommand{\softmax}{\mathrm{softmax}}

\begin{document}
%\doublespacing

\begin{flushright}
%INR-TH-
\end{flushright}
\title{Neural Network Quantum Field Theory from Transformer Architectures}

\author{Dmitry S. Ageev$^{a,b}$ and Yulia A. Ageeva$^{b,c,d}$}

\affiliation{$^{a}$Steklov Mathematical Institute, Russian Academy of Sciences,\\
Gubkin str. 8, 119991 Moscow, Russia\\
  $^{b}$Institute for Theoretical and Mathematical Physics,
  M.V.~Lomonosov Moscow State University, Leninskie Gory 1,
119991 Moscow,
Russia\\
$^{c}$Institute for Nuclear Research of
         the Russian Academy of Sciences,  60th October Anniversary
  Prospect, 7a, 117312 Moscow, Russia\\
 $^{d}$Department of Particle Physics and Cosmology, Physics Faculty, M.V. Lomonosov Moscow State University, Leninskie Gory 1-2, 119991 Moscow, Russia}

\emailAdd{ageev@mi-ras.ru}
\emailAdd{ageeva@inr.ac.ru}

\abstract{We propose a neural-network construction of Euclidean scalar quantum field theories from transformer attention heads, defining $n$-point correlators by averaging over random network parameters in the NN-QFT framework. For a single attention head, shared random softmax weights couple different width coordinates and induce non-Gaussian field statistics that persist in the infinite-width limit $d_k\to\infty$. We compute the two-point function in an attention-weight representation and show how Euclidean-invariant kernels can be engineered via random-feature token embeddings. We then analyze the connected four-point function and identify an ``independence-breaking'' contribution, expressible as a covariance over query--key weights, which remains finite at infinite width. Finally, we show that summing many independent heads with standard $1/N_h$ normalization suppresses connected non-Gaussian correlators as $1/N_h$, yielding a Gaussian NN-QFT in the large-head limit.
}

\maketitle

\newpage

% ============================================================
\section{Introduction}

Quantum field theory (QFT) is naturally characterized by its correlation functions, or equivalently by a measure over field configurations whose products reproduce the $n$-point functions. A complementary viewpoint has emerged from the large-width limit of neural networks: for many randomly initialized architectures, the induced distribution over functions converges to a Gaussian process (NNGP), with training in the same limit described by an associated neural tangent kernel (NTK), see, e.g. \cite{Lee:2017,Jacot:2018,Neal:1996,Rasmussen:2006}. In this regime the induced measure is Gaussian, so all connected correlators beyond the two-point function vanish and the corresponding field theory is ``free''. The neural network/field theory (NN-QFT) correspondence of \cite{Ferko:2026axm, Halverson:2021} sharpens this analogy by defining fields directly as random-network readouts and defining correlators by averaging over a chosen parameter ensemble. In that framework, Gaussian theories arise in the infinite-width limit when the constituent units are independently distributed, via the central limit theorem, while non-Gaussianity (``interactions'') can arise either from finite-width effects or from violations of independence that survive at infinite width \cite{Halverson:2021, Demirtas:2023fir}.

Transformers have become the standard architecture for sequence modeling and modern large language models, with attention \textit{heads} as the basic computational building block \cite{Bahdanau:2014,Vaswani:2017,JurafskyMartin:2026,Sakai:2026,Bordelon:2024,joshi2025transformers}. A single attention head mixes token representations through weights obtained by a softmax of query--key scores, producing a data-dependent linear combination of value vectors. The large-width behavior of attention networks has been investigated from the NNGP/NTK perspective \cite{Hron:2020}, where it is emphasized that the standard $\sqrt{d_k}$ (with $d_k$ being attention-head dimension) score normalization does not automatically enforce Gaussian asymptotics of attention outputs in the same way as in fully connected layers: the randomness of the softmax weights can persist and obstruct self-averaging \cite{Hron:2020}.

The main observation of this paper is that a \textit{single-head} transformer provides a particularly simple and intrinsic mechanism for ``independence breaking'' in the NN-QFT sense \cite{Halverson:2021,Demirtas:2023fir}. Different output coordinates of the head share the \textit{same} random attention weights, since those weights depend only on the query and key projections and are reused across all value dimensions. When we build a scalar field $\phi(x)$ as a linear readout of the head coordinates and define correlators by averaging over a Gaussian initialization ensemble, this shared-softmax structure couples the would-be independent width degrees of freedom and generates non-Gaussianity that survives the infinite-width limit $d_k\to\infty$.

This is of particular interest in the context of the NNGP/QFT correspondence. The presence of non-Gaussian behavior even in the infinite-width limit opens up a new possibility to naturally construct interacting QFTs from neural network architectures. Throughout this paper we use ``interacting'' in the minimal probabilistic sense: the field is non-Gaussian, equivalently some connected $n\ge 4$ correlators are nonzero. This notion is weaker than dynamical interaction in the standard QFT sense, but it is the appropriate notion for diagnosing departures from Gaussianity directly at the level of the induced measure over field configurations \cite{Halverson:2021,Demirtas:2023fir}.

After setting up the construction, we first compute the two-point function and emphasize the role of the embedding $x\mapsto \mathbf{x}(x)$ that represents spacetime points as tokens. As in \cite{Halverson:2021}, random Fourier feature embeddings provide a transparent way to engineer Euclidean-invariant kernels with tunable spectral density \cite{Rahimi:2007}. We then turn to the connected four-point function and show that its ``independence breaking'' contribution can be written as a covariance over the query--key weights, making manifest why it generically remains finite as $d_k\to\infty$. Finally, we extend the construction to many independent heads and show that, with the standard $1/N_h$ variance normalization, connected non-Gaussian correlators are suppressed as $1/N_h$ and vanish as $N_h\to\infty$.

The paper is organized as follows. In Section~2 we define the scalar field from a single attention head and compute its two-point function, including a Euclidean-invariant realization via random feature tokens. In Section~3 we compute the connected four-point function and isolate the single-head non-Gaussianity arising from shared attention weights. In Section~4 we study the multi-head generalization and show Gaussianization as $N_h\to\infty$.

% ============================================================
\section{Scalar field from single-head transformer}

For $x\in\mathbb{R}^d$ (the spacetime dimension is denoted by $d$) we follow \cite{Halverson:2021} and define the $n$-point correlation functions as
\begin{align}
G^{(n)}(x_1,\ldots,x_n)
&=\mathbb{E}_\theta\!\left[\phi(x_1)\cdots\phi(x_n)\right]
=\frac{1}{Z_\theta}\int d\theta\; \phi(x_1)\cdots\phi(x_n)\,P(\theta),
\label{eq:Gn_def_paper}
\end{align}
where
\begin{align}
    Z_\theta =\int d\theta\;P(\theta),
\end{align}
with $\theta$ being the full set of parameters of neural network (and $\phi$ depends on these parameters through the chosen architecture).
We introduce a scalar field as 
\begin{align}
\phi(x)
&=\sum_{i=1}^{d_k}\mathbf{z}_i\,\mathbf{head}_i(x),
\label{eq:phi_single_head_def_paper}
\end{align}
where $\mathbf{head}_i(x)$ denotes the $i$-th component of the transformer's head output for the token associated to the point $x$. The parameters $\mathbf{z}_i$ are drawn identically and independent distributed (i.i.d.). The attention-head dimension is denoted by $d_k$. In our conventions the $\mathbf{head}_i$ is a vector in $\mathbb{R}^{d_k}$. The index $i\in\{1,\ldots,d_k\}$ labels an output coordinate of a head. In the present construction, $d_k$ plays the role of a width parameter.

A single attention $\mathbf{head}_a(x)$ acting on token vectors $\mathbf{x}_a\in\mathbb{R}^d$\footnote{Token indices are written as $a,b,\ldots$ and label the input sequence elements $\mathbf{x}_a$. When we will compute correlators $G^{(n)}(x_1,\ldots,x_n)$ we will feed the corresponding tokens into attention; in the simplest presentation one may think of $\mathbf{x}_a=\mathbf{x}(x_a)$ as the token associated to the spacetime point $x_a$. We will later discuss the freedom in choosing the map $x\mapsto \mathbf{x}(x)$.} and can be formed as \cite{Vaswani:2017,JurafskyMartin:2026}:
\begin{align}
\mathbf{head}_a(x)
&=\sum^d_{b=1}\alpha_{ab}(x)\,\mathbf{v}_b (x).
\label{eq:head_def_paper}
\end{align}
The standard definitions of \textit{value, query, and key} vectors are
\begin{align}
\mathbf{v}_a=\mathbf{x}_a\,\mathbf{W}^{\mathbf{V}},
\qquad
\mathbf{q}_a=\mathbf{x}_a\,\mathbf{W}^{\mathbf{Q}},
\qquad
\mathbf{k}_a=\mathbf{x}_a\,\mathbf{W}^{\mathbf{K}},
\label{eq:qkv_defs_paper}
\end{align}
with the projection matrices $
\mathbf{W}^{\mathbf{Q}}\in\mathbb{R}^{d\times d_k}$, $\mathbf{W}^{\mathbf{K}}\in\mathbb{R}^{d\times d_k}$,
$\mathbf{W}^{\mathbf{V}}\in\mathbb{R}^{d\times d_k}$.
The \textit{attention weights} $\alpha_{ab}$ is
\begin{align}
\alpha_{ab}
=\softmax\!\left(\operatorname{score}(\mathbf{x}_a,\mathbf{x}_b)\right), \quad\operatorname{score}(\mathbf{x}_a,\mathbf{x}_b)
&\equiv \frac{\mathbf{q}_a\cdot\mathbf{k}_b}{\sqrt{d_k}},
\label{eq:score_alpha_paper}
\end{align}
where the softmax function converts a finite sequence (or ordered list) of numbers into a probability distribution over this sequence possible outcomes. Moreover, the result of a product $\mathbf{q}_a\cdot\mathbf{k}_b$ can be an arbitrarily large (positive or negative) value. The
exponentiating large values can lead to numerical issues, etc. That is why, one usually divide the dot product $\mathbf{q}_a\cdot\mathbf{k}_b$ by the square root of the dimensionality of the query and
key vectors (i.e. by $d_k$).

The crucial structural point is that the \textit{same} attention weights $\alpha_{ab}$ appear in every component $\mathbf{head}_i$: different output coordinates are coupled through the shared random softmax weights. This is exactly the place where the non-gaussian behaviour of single-head transformer comes from.
Finally, the full parameter set is
\begin{align}
\theta=\{\mathbf{z}_i,\theta_h\},
\quad
\theta_h=\{\mathbf{W}^{\mathbf{Q}},\mathbf{W}^{\mathbf{K}},\mathbf{W}^{\mathbf{V}}\}.
\end{align}

Let us take Gaussian initialization with a scaling adapted to fixed spacetime dimension $d$ as:
\begin{align}
\mathbb{E}[(\mathbf{W}^{\mathbf{Q}})^2]=\frac{\sigma_Q^2}{d},
\qquad
\mathbb{E}[(\mathbf{W}^{\mathbf{K}})^2]=\frac{\sigma_K^2}{d},
\qquad
\mathbb{E}[(\mathbf{W}^{\mathbf{V}})^2]=\frac{\sigma_V^2}{d},
\label{eq:W_scaling_paper}
\end{align}
with all matrix entries i.i.d.\ and these three matrices are independent. As it was mentioned above, the parameters $\mathbf{z}_i$ are i.i.d., so we consider
\begin{align}
\mathbb{E}[\mathbf{z}_i^{2n+1}]=0,
\qquad
\mathbb{E}[\mathbf{z}_i^2]=\frac{\sigma_z^2}{d_k},
\qquad
\mathbb{E}[\mathbf{z}_i^4]=\frac{\gamma_z^4}{d_k^2},
\label{eq:z_moments_paper}
\end{align}
where $\mathbb{E}[\mathbf{z}_i^{2n+1}]=0$ provides the $\mathbb{Z}_2$ symmetry, i.e.  $\phi \to - \phi$ for our real scalar field. 

For the further purposes we also note, that the large-width limit in this single-head setting is
\begin{align}
d_k\to\infty, 
\label{eq:large_dk_statement}
\end{align}
with $d$ fixed and with the variance scalings in
\eqref{eq:W_scaling_paper} --\eqref{eq:z_moments_paper} held fixed.

We assume the simplest model where all odd correlators
vanish, and begin with the two-point correlation function $G^{(2)}(x_1,x_2)$, i.e. \eqref{eq:Gn_def_paper} with $n=2$, which for the single-head field \eqref{eq:phi_single_head_def_paper} is
\begin{align}
G^{(2)}(x_1,x_2)
&=\sum_{i,j=1}^{d_k}\mathbb{E}[\mathbf{z}_i\mathbf{z}_j]\;
  \mathbb{E}\!\left[\mathbf{head}_i(x_1)\mathbf{head}_j(x_2)\right],
\label{eq:G2_expand_paper}
\end{align}
where we use
\begin{align}
\phi(x_1)\phi(x_2)
&=\left(\sum_{i=1}^{d_k}\mathbf{z}_i\,\mathbf{head}_i(x_1)\right)
  \left(\sum_{j=1}^{d_k}\mathbf{z}_j\,\mathbf{head}_j(x_2)\right)
=\sum_{i,j=1}^{d_k}\mathbf{z}_i\mathbf{z}_j\,
  \mathbf{head}_i(x_1)\mathbf{head}_j(x_2).
\end{align}
From \eqref{eq:z_moments_paper} we have
\begin{align}
\mathbb{E}[\mathbf{z}_i\mathbf{z}_j]=\delta_{ij}\frac{\sigma_z^2}{d_k},
\end{align}
so \eqref{eq:G2_expand_paper} becomes
\begin{align}
G^{(2)}(x_1,x_2)
&=\frac{\sigma_z^2}{d_k}\sum_{i=1}^{d_k} \mathbb{E}\!\left[\mathbf{head}_i(x_1)\mathbf{head}_i(x_2)\right].
\label{eq:G2_sum_over_i_paper}
\end{align}
Under the assumption that different output coordinates $i=1,\ldots,d_k$ are identically distributed at initialization, we may write
\begin{align}
\mathbb{E}\!\left[\mathbf{head}_i(x_1)\mathbf{head}_i(x_2)\right]
&\equiv H^{(2)}_{ii}(x_1,x_2),
\end{align}
with no sum on the right-hand side and $\mathbb{E}$ is averaging over the full set of parameters we have;
then \eqref{eq:G2_sum_over_i_paper} simplifies to
\begin{align}
G^{(2)}(x_1,x_2)
&=\sigma_z^2\,H^{(2)}_{ii}(x_1,x_2).
\label{eq:G2_final_paper}
\end{align}
In particular, there is no explicit $d_k$-dependence in $G^{(2)}(x_1,x_2)$ once the variance scaling in \eqref{eq:z_moments_paper} is chosen.

It is also useful to express $H^{(2)}_{ii}$ directly in terms of attention weights. Writing out the head definition \eqref{eq:head_def_paper} for the tokens corresponding to $x_1$ and $x_2$ gives
\begin{align}
\mathbf{head}_i(x_1)
&=\sum^d_{a= 1}\alpha_{ia}(x_1)\,\mathbf{v}_a(x_1),
\qquad
\mathbf{head}_i(x_2)
=\sum^d_{b = 1}\alpha_{ib}(x_2)\,\mathbf{v}_b(x_2),
\end{align}
hence
\begin{align}
\mathbf{head}_i(x_1)\mathbf{head}_i(x_2)
&=\sum^d_{a= 1}\sum^d_{b=1}
  \alpha_{ia}(x_1)\alpha_{ib}(x_2)\,\mathbf{v}_a(x_1)\mathbf{v}_b(x_2).
\label{eq:head_product_expand_paper}
\end{align}
Averaging over $\mathbf{W}^{\mathbf{V}}$ with \eqref{eq:W_scaling_paper} gives
\begin{align}
\mathbb{E}_{\mathbf{W}^{\mathbf{V}}}\!\left[\mathbf{v}_a\mathbf{v}_b\right]
&=\frac{\sigma_V^2}{d}\,(\mathbf{x}^{(1)}_a\cdot\mathbf{x}^{(2)}_b),
\label{eq:vv_average_paper}
\end{align}
where the token $\mathbf{x}^{(1)}_{a}$ relates to the $x_1$ coordinate and
where we also use \eqref{eq:qkv_defs_paper}.
Combining \eqref{eq:head_product_expand_paper} and \eqref{eq:vv_average_paper} we obtain the explicit representation
\begin{align}
H^{(2)}_{ii}(x_1,x_2)
&=\frac{\sigma_V^2}{d}\;
\mathbb{E}_{\mathbf{W}^{\mathbf{Q}},\mathbf{W}^{\mathbf{K}}}\!\left[
\sum^d_{a= 1}\sum^d_{b=1}\alpha_{ia}(x_1)\alpha_{ib}(x_2)\,(\mathbf{x}^{(1)}_{a}\cdot\mathbf{x}^{(2)}_{b})
\right],
\label{eq:H2_attention_form_paper}
\end{align}
and therefore
\begin{align}
G^{(2)}(x_1,x_2)
&=\sigma_z^2\,\frac{\sigma_V^2}{d}\;
\mathbb{E}_{\mathbf{W}^{\mathbf{Q}},\mathbf{W}^{\mathbf{K}}}\!\left[
\sum^d_{a= 1}\sum^d_{b=1}\alpha_{ia}(x_1)\alpha_{ib}(x_2)\,(\mathbf{x}^{(1)}_{a}\cdot\mathbf{x}^{(2)}_{b})
\right].
\label{eq:G2_attention_form_paper}
\end{align}
This formula exhibits clearly where the transformer structure enters: all nontrivial dependence beyond the value inner products $(\mathbf{x}^{(1)}_{a}\cdot\mathbf{x}^{(2)}_{b})$ is carried also by the softmax weights $\alpha$.

A distinctive feature of the neuron construction in \cite{Halverson:2021} is the freedom to choose the functional form of the neurons. In the transformer setting, the attention map \eqref{eq:head_def_paper} fixes the architecture, but one retains substantial freedom in the choice of the embedding $x\mapsto \mathbf{x}(x)$, i.e.\ how a spacetime point is represented as a token. One particularly transparent choice uses random Fourier features \cite{Rahimi:2007}, for example the \textit{cos-net} ensembles \cite{Halverson:2021} with token components are given by:
\begin{align}
\mathbf{x}_r(x)
&=\sqrt{2}\,F(\mathbf{b}_r)\,
  \cos\!\Big(\sum_{s=1}^{d} b_{rs}x_s+c_r\Big),
\qquad
r=1,\ldots,d,
\label{eq:random_feature_token_paper}
\end{align}
where $F(\mathbf{b}_r)$ is an amplitude profile and $(b_{ab},c_a)$ being the random parameters and $\mathbf{b}_a$ is just a $i$-th row of matrix $b$ (here we follow the notations of \cite{Halverson:2021}). 
It turns out that such a construction respects the Euclidean invariance what is crucial for realistic QFT. In this construction the random feature parameters $(\mathbf{b}_r,c_r)$ are treated as additional part of $\theta_h$ set and are averaged over in correlators too. The tokens $\mathbf{x}(x)$ are therefore random functions of $x$ under $\mathbb{E}_{\theta_h}$, even if $x$ itself is deterministic.

To see why this is useful, consider the simplest regime in which attention does not mix different tokens, so that the head output reduces effectively to the value,
\begin{align}
\mathbf{head}_i(x)=\mathbf{v}_i(x)=\mathbf{x}_i(x)\,\mathbf{W}^{\mathbf{V}}.
\label{eq:self_attention_regime_paper}
\end{align}
Then \eqref{eq:vv_average_paper} gives
\begin{align}
H^{(2)}_{ii}(x_1,x_2)
&=\mathbb{E}_{\theta_h}\!\left[\mathbf{head}_i(x_1)\mathbf{head}_i(x_2)\right]
=\frac{\sigma_V^2}{d}\,
  \mathbb{E}_{\mathbf{b},c}\!\left[\mathbf{x}_i(x_1)\cdot\mathbf{x}_i(x_2)\right].
\label{eq:H2_dotprod_regime_paper}
\end{align}
Now compute the dot product average using \eqref{eq:random_feature_token_paper}. For each $i$ we have
\begin{align}
\mathbf{x}_i(x_1)\mathbf{x}_i(x_2)
&=2F(\mathbf{b}_r)^2\,
  \cos\!\Big(\sum_s b_{is}(x_1)_s+c_i\Big)\,
  \cos\!\Big(\sum_k b_{ik}(x_2)_k+c_i\Big),
\end{align}
and averaging over a phase $c_i$ assuming that $c \sim U[-\pi,\pi]$ yields
\begin{align}
\mathbb{E}_{c_i}\!\left[\mathbf{x}_i(x_1)\mathbf{x}_i(x_2)\right]
&=F(\mathbf{b}_i)^2\,
  \cos\!\Big(\sum_s b_{is}\big((x_1)_s-(x_2)_s\big)\Big).
\end{align}
Averaging over i.i.d.\ $\mathbf{b}_i$ gives a translation-invariant kernel depending only on $x_1-x_2$:
\begin{align}
\mathbb{E}_{\mathbf{b},c}\!\left[\mathbf{x}_i(x_1)\cdot\mathbf{x}_i(x_2)\right]
&=\mathbb{E}_{\mathbf{b}}\!\left[
F(\mathbf{b}_i)^2\,
  \cos\!\Big(\sum_s b_{is}\big((x_1)_s-(x_2)_s\big)\Big)
\right].
\label{eq:dotprod_random_features_paper}
\end{align}
Combining \eqref{eq:G2_final_paper}, \eqref{eq:H2_dotprod_regime_paper}, and \eqref{eq:dotprod_random_features_paper} yields
\begin{align}
G^{(2)}(x_1,x_2)
&=\sigma_z^2\,\sigma_V^2\;
\mathbb{E}_{\mathbf{b}}\!\left[
F(\mathbf{b}_i)^2\,
  \cos\!\Big(\sum_s b_{is}\big((x_1)_s-(x_2)_s\big)\Big)
\right].
\label{eq:G2_random_features_paper}
\end{align}
This expression should be compared with the standard free Euclidean propagator,
\begin{align}
G^{(2)}_{\mathrm{free}}(x_1,x_2)
&=\int\frac{d^dp}{(2\pi)^d}\;
\frac{\cos\!\Big(\sum_s p_s\big((x_1)_s-(x_2)_s\big)\Big)}{p^2+m^2},
\label{eq:free_propagator_cos_form_paper}
\end{align}
where we write the cosine function instead of standard exponent.
By choosing the distribution of $\mathbf{b}$ and the profile $F(\mathbf{b})$ so that the effective spectral density matches $(p^2+m^2)^{-1}$ (up to the usual regularization subtleties), one can engineer $G^{(2)}(x_1,x_2)$ to coincide with \eqref{eq:free_propagator_cos_form_paper}. In this sense the freedom in the token map $x\mapsto \mathbf{x}(x)$ plays the role of choosing a neuron ensemble in \cite{Halverson:2021}. Importantly, none of this alters the transformer architecture; it only specifies the input representation and the parameter ensemble over which we average. Our simple example \eqref{eq:self_attention_regime_paper} fully covers the model with ``cos-net neurons'' from \cite{Halverson:2021}, thus also respects the translation and rotational invariance as well as satisfies the
Osterwalder-Schrader (OS) axioms \cite{Osterwalder:1973,Osterwalder:1975}. Moreover, if we consider the general formula for head \eqref{eq:head_def_paper} with attention weights $\alpha_{ij}$, the translation and rotational invariance remains within cos-net choice \eqref{eq:random_feature_token_paper} for tokens.

\section{Four-point function and non-Gaussianity from shared attention weights within single-head transformer}

The construction of transformer's single head attention \eqref{eq:head_def_paper} admits the non-gaussian behavior, see \cite{Demirtas:2023fir,Hron:2020,Sakai:2026}, even in the $d_k \to \infty$ limit. This property is very useful since  gives rise to different correlation functions with $n>2$ when constructing QFT with interaction. We thus turn to the connected four-point function, the simplest diagnostic of non-Gaussianity. Using \eqref{eq:phi_single_head_def_paper}, the four-point function expands as
\begin{align}
\phi(x_1)\phi(x_2)\phi(x_3)\phi(x_4)
&=\sum_{i,j,k,\ell=1}^{d_k}
\mathbf{z}_i\mathbf{z}_j\mathbf{z}_k\mathbf{z}_\ell\;
\mathbf{head}_i(x_1)\mathbf{head}_j(x_2)\mathbf{head}_k(x_3)\mathbf{head}_\ell(x_4).
\end{align}
Taking expectation and using \eqref{eq:z_moments_paper}, as well as following the notations of \cite{Halverson:2021}, we write the connected part as
\begin{align}
G^{(4)}_c(x_1,x_2,x_3,x_4)
&= I^{(4)}_{d_k}(x_1,x_2,x_3,x_4)
  + I^{(4)}_{\mathrm{IB}}(x_1,x_2,x_3,x_4),
\label{eq:G4c_split_paper}
\end{align}
where ``IB'' means independence breaking 
and these two contributions can be expressed (with no sums implied on the right-hand sides) as
\begin{align}
I^{(4)}_{d_k}
&=\frac{\gamma_z^4}{d_k}\,H_{iiii}
  -\frac{\sigma_z^4}{d_k}\Big(
    H^{12}_{ii}H^{34}_{ii}
   +H^{13}_{ii}H^{24}_{ii}
   +H^{14}_{ii}H^{23}_{ii}
  \Big),
\label{eq:Idk4_paper}
\\
I^{(4)}_{\mathrm{IB}}
&=\sigma_z^4(1-\delta_{ij})\Big(1-\frac{1}{d_k}\Big)
  \Big[
   \big(H_{iijj}-H^{12}_{ii}H^{34}_{jj}\big)
  +\big(H_{ijij}-H^{13}_{ii}H^{24}_{jj}\big)
  +\big(H_{ijji}-H^{14}_{ii}H^{23}_{jj}\big)
  \Big],
\label{eq:IIB4_paper}
\end{align}
where
\begin{align}
H^{ab}_{ii}&\equiv H^{(2)}_{ii}(x_a,x_b),
\qquad
H_{iijj}\equiv H^{(4)}_{iijj}(x_1,x_2,x_3,x_4),
\end{align}
and similarly for $H_{ijij}$ and $H_{ijji}$. The term $I^{(4)}_{d_k}$ is a genuine finite-width effect (to obtain non-gaussianities) and scales as $1/d_k$. In contrast, $I^{(4)}_{\mathrm{IB}}$ vanishes if different output coordinates are independent in the sense that mixed correlators factorize, but it can remain finite as $d_k\to\infty$ when that independence is broken. The latter is precisely what transformer's weights $\alpha_{ab}$ provides.

The mechanism underlying $I^{(4)}_{\mathrm{IB}}\neq 0$ in the transformer setting is conceptually simple: different output coordinates share the same random attention weights $\alpha_{ab}$. It is useful to see this explicitly for the following terms from \eqref{eq:IIB4_paper}, for $i\neq j$:
\begin{align}
H_{iijj}-H^{12}_{ii}H^{34}_{jj}.
\label{eq:IB_bracket_paper}
\end{align}
We split the head parameters as $\theta_h=\{\mathbf{W}^{\mathbf{Q}},\mathbf{W}^{\mathbf{K}},\mathbf{W}^{\mathbf{V}}\}$ and condition on $(\mathbf{W}^{\mathbf{Q}},\mathbf{W}^{\mathbf{K}})$. Conditional on these, the weights $\alpha_{ab}$ are deterministic numbers, and the $i$-th and $j$-th components of $\mathbf{head}$ depend on independent columns of $\mathbf{W}^{\mathbf{V}}$. As a result,
\begin{align}
\mathbb{E}_{\mathbf{W}^{\mathbf{V}}}\!\left[
\mathbf{head}_i(x_1)\mathbf{head}_i(x_2)\mathbf{head}_j(x_3)\mathbf{head}_j(x_4)
\ \Big|\ \mathbf{W}^{\mathbf{Q}},\mathbf{W}^{\mathbf{K}}
\right]
&=
X_{12}\,X_{34},
\label{eq:conditional_factorization_paper}
\end{align}
where we have introduced the conditional two-point objects
\begin{align}
X_{ab}
&\equiv \mathbb{E}_{\mathbf{W}^{\mathbf{V}}}\!\left[
\mathbf{head}_i(x_a)\mathbf{head}_i(x_b)
\ \Big|\ \mathbf{W}^{\mathbf{Q}},\mathbf{W}^{\mathbf{K}}
\right].
\label{eq:Xab_def_paper}
\end{align}
Taking expectation over $(\mathbf{W}^{\mathbf{Q}},\mathbf{W}^{\mathbf{K}})$ now gives
\begin{align}
H_{iijj}
&=\mathbb{E}_{\mathbf{W}^{\mathbf{Q}},\mathbf{W}^{\mathbf{K}}}\!\left[X_{12}X_{34}\right],
\qquad
H^{12}_{ii}H^{34}_{jj}
=\mathbb{E}_{\mathbf{W}^{\mathbf{Q}},\mathbf{W}^{\mathbf{K}}}[X_{12}]\;
 \mathbb{E}_{\mathbf{W}^{\mathbf{Q}},\mathbf{W}^{\mathbf{K}}}[X_{34}],
\end{align}
and therefore
\begin{align}
H_{iijj}-H^{12}_{ii}H^{34}_{jj}
&=\mathbb{E}_{\mathbf{W}^{\mathbf{Q}},\mathbf{W}^{\mathbf{K}}}\!\left[X_{12}X_{34}\right]
 -\mathbb{E}_{\mathbf{W}^{\mathbf{Q}},\mathbf{W}^{\mathbf{K}}}[X_{12}]\;
  \mathbb{E}_{\mathbf{W}^{\mathbf{Q}},\mathbf{W}^{\mathbf{K}}}[X_{34}].
\label{eq:covariance_form_paper}
\end{align}
Thus the bracket \eqref{eq:IB_bracket_paper} is a covariance over $(\mathbf{W}^{\mathbf{Q}},\mathbf{W}^{\mathbf{K}})$. It vanishes only if $X_{ab}$ is (essentially) non-random as a function of these weights.

The $X_{ab}$ can be written explicitly by repeating the $\mathbf{W}^{\mathbf{V}}$-average performed in \eqref{eq:H2_attention_form_paper}. One finds
\begin{align}
X_{ab}
&=\frac{\sigma_V^2}{d}\sum_{u}\sum_{v}\alpha_{au}\alpha_{bv}\,(\mathbf{x}_u\cdot\mathbf{x}_v),
\label{eq:Xab_attention_paper}
\end{align}
which is manifestly a nontrivial random functional of the softmax weights. In particular, choosing $(x_3,x_4)=(x_1,x_2)$ turns \eqref{eq:covariance_form_paper} into a variance,
\begin{align}
H_{iijj}(x_1,x_2,x_1,x_2)-H^{12}_{ii}H^{12}_{jj}
&=\mathrm{Var}_{\mathbf{W}^{\mathbf{Q}},\mathbf{W}^{\mathbf{K}}}(X_{12}),
\end{align}
where $\mathrm{Var}(X) =\mathbb{E}\left[X^2\right]-\mathbb{E}[X]^2$ and
which is strictly positive unless $X_{12}$ is almost surely constant. The variance of the score from \eqref{eq:score_alpha_paper} shows that the attention weights $\alpha$ do not freeze in the large-$d_k$ limit with the scaling \eqref{eq:W_scaling_paper}; hence the variance is generically nonzero. We conclude that $I^{(4)}_{\mathrm{IB}}$ is generically finite as $d_k\to\infty$, and the single-head field is interacting in the minimal probabilistic sense stated above.

\section{Many heads and Gaussian NN-QFT}

As we have seen, the non-gaussianies arise in the single-head setup \cite{Demirtas:2023fir,Hron:2020}.\footnote{The non-gaussian behavior can be also ``killed'' changing the normalization of the score \eqref{eq:score_alpha_paper}: in \cite{Hron:2020,Sakai:2026,Yang:2022} it was shown, that dividing by $d_k$ (not the $\sqrt{d_k}$ as in \eqref{eq:score_alpha_paper}) makes model gaussian at $d_k\to\infty$ limit.} We now extends the construction to $N_h$ attention heads (multi-heads) \cite{Bordelon:2024} and explicitly show that this many-head setup is actually a Gaussian NN-QFT. For each head $h=1,\ldots,N_h$ we introduce independent weight matrices $\mathbf{W}^{\mathbf{Q}}_{h}, \mathbf{W}^{\mathbf{K}}_{h},\mathbf{W}^{\mathbf{V}}_{h}$, 
each distributed as in \eqref{eq:W_scaling_paper}, and we define head outputs $\mathbf{head}^{(h)}(x)$ by the same attention formulas as before, with $\mathbf{W}^{\mathbf{Q}},\mathbf{W}^{\mathbf{K}},\mathbf{W}^{\mathbf{V}}$ replaced by $\mathbf{W}^{\mathbf{Q}}_{h},\mathbf{W}^{\mathbf{K}}_{h},\mathbf{W}^{\mathbf{V}}_{h}$. The scalar field is taken to be the sum of readouts of each head,
\begin{align}
\phi(x)
&=\sum_{h=1}^{N_h}\sum_{i=1}^{d_k}\mathbf{z}_{hi}\,\mathbf{head}^{(h)}_i(x).
\label{eq:phi_multihead_def_paper}
\end{align}
To keep $G^{(2)}$ finite as $N_h$ varies, the appropriate scaling of the readout weights is now
\begin{align}
\mathbb{E}[\mathbf{z}_{hi}^2]
&=\frac{\sigma_z^2}{N_h\,d_k},
\qquad
\mathbb{E}[\mathbf{z}_{hi}^4]
=\frac{\gamma_z^4}{N_h^2\,d_k^2},
\label{eq:z_multihead_scaling_paper}
\end{align}
with all $\mathbf{z}_{hi}$ i.i.d.\ and independent of all head weights.

The two-point function is computed exactly as in the single-head case: the Kronecker deltas collapse the sums over $(h,i)$ and yield
\begin{align}
G^{(2)}(x_1,x_2)
&=\sigma_z^2\,H^{(2)}_{ii}(x_1,x_2),
\end{align}
so $G^{(2)}$ is independent of $N_h$ when the scaling \eqref{eq:z_multihead_scaling_paper} is used.

The behavior of non-Gaussianity is very different. Define the contribution of a single head to the scalar field by
\begin{align}
\phi^{(h)}(x)
& = \sum_{i=1}^{d_k}\mathbf{z}_{hi}\,\mathbf{head}^{(h)}_i(x),
\qquad
\phi(x)=\sum_{h=1}^{N_h}\phi^{(h)}(x).
\end{align}
By construction, different heads are independent at initialization. A standard cumulant computation then shows that the connected four-point function of the sum is the sum of the connected four-point functions of the individual heads:
\begin{align}
G^{(4)}_c(x_1,x_2,x_3,x_4)
&=\sum_{h=1}^{N_h}G^{(4)}_{c,h}(x_1,x_2,x_3,x_4),
\label{eq:Gc4_sum_over_heads_paper}
\end{align}
where $G^{(4)}_{c,h}$ denotes the connected four-point function of $\phi^{(h)}$.

Now the key point is scaling: for each head, the formulas \eqref{eq:Idk4_paper}, \eqref{eq:IIB4_paper} apply with $d_k$ as width, but with $\sigma_z^2$ and $\gamma_z^4$ effectively replaced by
\begin{align}
\sigma_z^2\to \frac{\sigma_z^2}{N_h},
\qquad
\gamma_z^4\to \frac{\gamma_z^4}{N_h^2},
\end{align}
because the moments of $\mathbf{z}_{hi}$ carry the $1/N_h$ factors in \eqref{eq:z_multihead_scaling_paper}. Consequently,
\begin{align}
G^{(4)}_{c,h}(x_1,x_2,x_3,x_4)
&=\mathcal{O}\!\left(\frac{1}{N_h^2}\right),
\end{align}
and summing over $h$ in \eqref{eq:Gc4_sum_over_heads_paper} yields the large-$N_h$ suppression
\begin{align}
G^{(4)}_c(x_1,x_2,x_3,x_4)
&=\mathcal{O}\!\left(\frac{1}{N_h}\right)
\qquad\Longrightarrow\qquad
\lim_{N_h\to\infty}G^{(4)}_c(x_1,x_2,x_3,x_4)=0.
\label{eq:multihead_gaussianization_paper}
\end{align}
Thus, when many independent heads are combined with the standard $1/N_h$ variance normalization, connected non-Gaussian correlators vanish in the large number-of-heads limit. In this sense multi-head averaging make setup gaussian, even though a single head can produce $I^{(4)}_{\mathrm{IB}}\neq 0$ at infinite $d_k$.

% ============================================================
\section{Conclusion}

We constructed a class of neural network field theories generated by transformer attention heads by defining a scalar field as a linear readout of a head output and defining Euclidean correlators by averaging over a Gaussian initialization ensemble, following the general NN-QFT perspective of \cite{Halverson:2021,Demirtas:2023fir}. The resulting two-point function admits a compact representation in terms of attention weights and value inner products, and by choosing the token map $x\mapsto\mathbf{x}(x)$ via random Fourier features one can engineer Euclidean-invariant kernels with tunable spectra in direct analogy with Euclidean-invariant neuron ensembles \cite{Halverson:2021,Rahimi:2007}. With appropriate choices, this reproduces standard free-field propagators while keeping the transformer architecture fixed.

Our main result is that, unlike standard independently distributed neuron ensembles, a \textit{single} attention head is generically non-Gaussian even at infinite width $d_k\to\infty$. The connected four-point function splits into a finite-width contribution that scales as $1/d_k$ and an ``independence breaking'' contribution associated with the shared attention weights. Because the same random softmax weights appear in every output coordinate, mixed correlators fail to factorize; the independence-breaking term can be written as a covariance over the query--key weights and is therefore nonzero whenever the attention map does not freeze \cite{Hron:2020}. In the minimal probabilistic sense adopted here, a single-head transformer thus induces an interacting (non-Gaussian) field theory without relying on finite-width corrections \cite{Halverson:2021,Demirtas:2023fir}.

We also showed that this mechanism is suppressed by averaging over many independent heads: with the standard $1/N_h$ normalization of readout variances, connected non-Gaussian correlators decay as $1/N_h$ and vanish as $N_h\to\infty$. Thus, while multi-head attention is central to practical transformer models \cite{Vaswani:2017,JurafskyMartin:2026}, the simplest transformer-induced non-Gaussian NN-QFTs arise already at the level of a single head. Extending these results beyond a single attention layer (including, for example, residual streams and different normalizations), as well as understanding how training deforms the induced measure and correlators and how to extract (effective) actions using the general machinery of \cite{Demirtas:2023fir} as well as how to correctly build, for example, $\phi^4$-interaction in the framework of single-head transformer are all natural directions for future work.

% ============================================================

\end{document}